# A Deeper Look at Dataset Bias

Tatiana Tommasi · Novi Patricia · Barbara Caputo · Tinne Tuytelaars

**Abstract** The presence of a bias in each image data collection has recently attracted a lot of attention in the computer vision community showing the limits in generalization of any learning method trained on a specific dataset. At the same time, with the rapid development of deep learning architectures, the activation values of Convolutional Neural Networks (CNN) are emerging as reliable and robust image descriptors. In this paper we propose to verify the potential of the DeCAF features when facing the dataset bias problem. We conduct a series of analyses looking at how existing datasets differ among each other and verifying the performance of existing debiasing methods under different representations. We learn important lessons on which part of the dataset bias problem can be considered solved and which open questions still need to be tackled.

## 1 Introduction

Since its spectacular success in the 2012 edition of the Imagenet Large Scale Visual Recognition Challenge (ILSVRC, [34]), deep learning has dramatically changed the research landscape in visual recognition [26]. By training a Convolutional Neural Network (CNN) over millions of data it is possible to get impressively high quality object annotations and detections. A large number of studies have recently proposed improvements over the CNN architecture of Krizhevsky *et al.* [26] with the aim to better suit an ever increasing typology of visual applications [21,45,35]. At the same time, the activation values of the final hidden layers have quickly gained the status

T. Tommasi
Department of Computer Science
University of North Carolina at Chapel Hill, Chapel Hill, NC, USA
E-mail: ttommasi@cs.unc.edu

N. Patricia
Idiap Research Institute, Martigny CH
EPFL, Lausanne, CH

B. Caputo
University of Rome La Sapienza
Rome, Italy

T. Tuytelaars
KU Leuven, ESAT, PSI, iMind
Belgium

of off-the-shelf state of the art features [33]. Indeed, several works demonstrated that DeCAF (as well as Caffe [7], Overfeat [37], VGG-CNN [4] and other specific implementations) can be used as powerful image descriptors [4,17]. The improvements obtained over previous methods are so impressive that one might wonder whether they can be considered as a sort of "universal features", *i.e.* image descriptors that can be helpful in any possible visual recognition problem.

The aim of this paper is to contribute to answering this question when focusing on dataset bias. Previous work seemed to imply that the generalization issue faced when training and testing on data extracted from different collections was solved, or on the way to be solved, by using such features [7,44]. However, the analysis is generally restricted to controlled cases where the data difference is limited to specific visual domain shift [7,23]. Some papers have also considered the supervised setting where labelled samples of the target dataset are available [44] or have investigated the possibility to modify by fine-tuning the CNN final output [32]. Although related to the dataset bias problem, this transfer learning scenario does not apply when the final goal is to test on data for which no annotations are available.

Our goal here is to make an in depth analysis of the dataset bias problem when using DeCAF features. To this end, we make two contributions:

1. we asses the performance of DeCAF features on the most comprehensive experimental setup existing for dataset bias. We build on the setting proposed in [38], consisting of a cross-dataset testbed over twelve different databases. We tailor this setting to the specificity of the dataset bias problem, and we run extensive experiments on it;
2. we propose a new measure to evaluate quantitatively the ability of a given algorithm to address the dataset bias. As opposed to what proposed previously in the literature [39], our measure takes into account both the performance obtained on the in-dataset task and the percentage drop in performance across datasets.

Our analysis explores the key aspects of dataset bias including cross-dataset generalization, how to undo the dataset bias, and up to which extent domain adaptation algorithms might help. The experiments highlight several important aspects of the problem and the suitability of DeCAF features for attacking it. In particular we found that (1) the negative bias persists; (2) attempts of undoing the dataset bias with existing ad-hoc learning algorithms do not help; (3) some previously discarded adaptive strategies appear extremely effective. The picture emerging from these findings is that of a problem open for research and in need for new directions, able to accommodate at the same time the potential of deep learning and the difficulties of large scale cross-database generalization.

## 2 What is Dataset Bias?

The visual world is so complex and nuanced that any finite set of samples ends up describing just some of its aspects. Moreover, in case the samples are collected for a particular task, they will inevitably cover just some specific visual region. Hence, it is not surprising that pre-defined image collections, like existing computer vision datasets, present such specific bias to be easily recognizable [39]. The main causes have been pointed out and named in [39]. The *capture bias* is related to how the images are acquired both in terms of the used device and of the collector preferences for point of view, lighting conditions, etc. The *category* or *label bias* comes from the fact that visual semantic categories are often poorly defined: similar images may be annotated with different names while, due to the in-class variability, the same name can be assigned to visually different images. Finally, each collection may contain a distinct set of categories and this causes the *negative bias*. If we focus only on the classes shared among them, the "rest of the world" will be defined differently depending on the collection.

Overall, the problem of *dataset bias* originates from the culprits listed above: the limited nature of existing datasets, created to evaluate learning models, might induce false conclusions. Apart from trying to create new and less biased data collections, the ultimate question that we need to answer is: **how can we use available data to generalize to new unseen samples even when training and test collections are different**? This dilemma goes beyond computer vision and it is a long standing topic in machine learning under the names of domain shift, sample selection bias and class imbalance [24]. Specifically, a *domain* is a set of data defined by its marginal and conditional distributions with respect to the assigned labels. The *sample selection bias* (or covariate shift) is due to a marginal distribution difference between two domains and mainly corresponds to what indicated above as capture bias. Intuitively, if we consider two collections containing the same set of categories, the bias remains mainly due to the chosen representation. Thus, it can be removed by using a feature that properly encodes the category information regardless of the dataset-specific exogenous factors. Several feature-adaptive methods have been proposed for this task [12, 16] and CNN descriptors demonstrated to be robust against this bias [7]. Still, the class imbalance causes a difference among the conditional distributions of two domains and it is related both to the category and negative bias mentioned above.

Measuring how two sets of data are related gives information about their respective bias. Theoretical studies [1] indicate that the error on unseen data across domains depends mainly on three terms. One is obviously the *training error*: a model that does not represent properly the training data cannot be effective on new samples. The second is the *domain divergence* which mainly depends on the marginal distribution shift *i.e.* on how much the two sets of data overlap among each other in the chosen feature space. The third depends on the difference among the respective conditional distributions of each collection and on the possibility to have a *model that fits both of them*. The *name the dataset* test used in [39] gives an indication of the similarity among the collections, but due to the different sets of covered classes, it is difficult to separate the effect of the marginal and of the conditional distribution shift or give insight on how well a model trained on one dataset will perform on another. The cross-dataset performance drop [39, 16] is also often used as generalization measure, but it disregards the importance of the training error on the source dataset.

Due to their intrinsic relation, the problem of dataset bias and domain adaptation have often be faced together. Several domain adaptive methods have been tested on setups extracted from multiple datasets [12, 16] and some approaches proposed to identify the latent domains in the data collections before applying domain adaptation solutions [22,14,43]. However, often the experimental tests are performed on a limited amount of image samples, or the methods are not suited to deal with the *big-data* issue. Considering the current dimensionality of data collections, the new state of the art established by the CNN features, and the growing interest towards models that should be able to generalize while learning on-the-fly [5], it is now time to make a point on the dataset bias problem with its old and new challenges.

## 3 Evaluation Protocol

We describe here the setup adopted for the experiments and we introduce the measures used to evaluate the cross-dataset generalization performance.

*Datasets & Features.* We focus on twelve datasets, created and used before for object categorization, that have been recently organized in a cross-dataset testbed with the definition of two data setups [38]:

- **sparse set**. It contains 105 Imagenet classes [6] aligned to 95 classes of Caltech256 [19] and Bing [40], 89 classes of SUN [42], 35 classes of Caltech101 [11], 17 classes of Office [36], 18 classes of RGB-D [27], 16 classes of Animals with Attributes (AwA) [28] and Pascal VOC07 [9], 13 classes of MSRCORID [31], 7 classes of ETH80 [29], and 4 classes of a-Yahoo [10].
- **dense set**. It contains 40 classes shared by Bing, Caltech256, Imagenet and SUN.

The testbed has been released[1] together with three feature representations:

- **BOWsift**: dense SIFT descriptors [30] extracted by following the same protocol previously used for the ILSVRC2010 contest[2] and quantized into a BOW representation based on a vocabulary of 1000 visual words;
- **DeCAF6, DeCAF7**: the mean-centered raw RGB pixel intensity values of all the collection images (warped to 256x256) are given as input to the CNN architecture of Krizhevsky *et al.* by using the DeCAF implementation[3]. The activation values of the 4096 neurons in the 6-th and 7-th layers of the network are considered as image descriptors.

In our experiments we use the L2-normalized version of the feature vectors and adopt the z-score normalization for the BOWsift features when testing domain adaptation methods. We mostly focus on the results obtained with the DeCAF features and use the BOWsift representation as a reference baseline.

*Evaluation Measures.* Our basic experimental setup considers both in-dataset and cross-dataset evaluations. With *in-dataset* we mean training and testing on samples extracted from the same dataset, while with *cross-dataset* we indicate experiments where training and testing samples belongs to different collections. We use $Self$ to specify the in-dataset performance and $Mean\ Other$ for the average cross-dataset performance over multiple test collections.

---

[1] https://sites.google.com/site/crossdataset/
[2] http://www.image-net.org/challenges/LSVRC/2010/
[3] https://github.com/UCB-ICSI-Vision-Group/decaf-release/

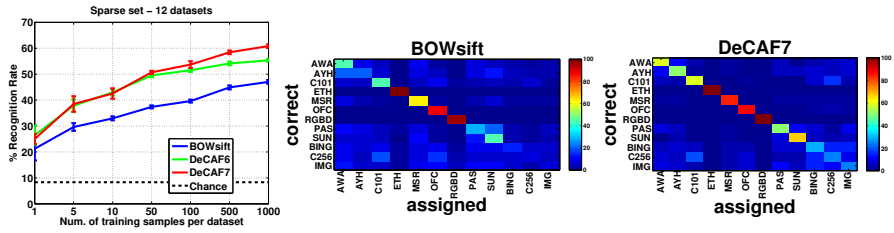

**Fig. 1** Name the dataset experiment over the sparse setup with 12 datasets. We use a linear SVM classifier with C value tuned by 5-fold cross validation on the training set. We show average and standard deviation results over 10 repetitions. The title of each confusion matrix indicates the feature used for the corresponding experiments.

One way to quantitatively evaluate the cross dataset generalization was previously proposed in [39]. It consists of measuring the percentage drop ($\% \, Drop$) between $Self$ and $Mean \, Others$. However, being a relative measure, it looses the information on the value of $Self$ which is important if we want to compare the effect of different learning methods or different representations. In fact a 75% drop w.r.t a 100% self average precision has a different meaning than a 75% drop w.r.t. a 25% self average precision. To overcome this drawback, we propose here a different *Cross-Dataset (CD)* measure defined as

$$CD = \frac{1}{1 + exp^{-\{(Self - Mean \; Others)/100\}}} \; .$$

$CD$ uses directly the difference ($Self - Mean \, Others$) while the sigmoid function rescales this value between 0 and 1. This allows for the comparison among the results of experiments with different setups. Specifically $CD$ values over $0.5$ indicate a presence of a bias, which becomes more significant as $CD$ gets close to 1. On the other hand, $CD$ values below $0.5$ correspond to cases where either $Mean \, Other \geq Self$ or the $Self$ result is very low. Both these conditions indicate that the learned model is not reliable on the data of its own collection and it is difficult to draw any conclusion from its cross-dataset performance.

## 4 Studying the Sparse set

4.1 Name the Dataset

With the aim of getting an initial idea on the relation among the datasets and how different representations capture their specific content, we start our analysis by running the *name the dataset* test on the sparse data setup. We extract randomly 1000 images from each of the 12 collections and we train a 12-way linear SVM classifier that we then test on a disjoint set of 300 images. The experiment is repeated 10 times with different data splits and we report the obtained average results in Figure 1. The plot on the left shows the recognition rate as a function of the training set size and indicates that DeCAF allows for a much better separation among the collections than what is obtained with BOWsift. In particular DeCAF7 shows an advantage over DeCAF6 for large number of training samples. From the confusion matrices (middle and right in Figure 1) we see that it is easy to distinguish ETH80, Office and RGB-D datasets from all the others regardless of the used representation, given the specific lab-nature of these collections. DeCAF captures better than BOWsift the characteristics of A-Yahoo, MSRCORID, Pascal VOC07 and SUN, improving the recognition results on them. Finally, Bing, Caltech256 and Imagenet are the datasets with the highest confusion level, an effect mainly due to the large number of classes and images per class. Still, this confusion decreases when using DeCAF.

| | BOWsift | % Drop | CD | DeCAF6 | % Drop | CD | DeCAF7 | % Drop | CD |
|---|---|---|---|---|---|---|---|---|---|
| Car | P: 28.6 26.4 26.7 29.2 | 3.9 | 0.50 | P: 26.2 19.2 44.3 42.6 | -35.1 | 0.47 | P: 14.5 11.8 28.9 28.3 | -59.1 | 0.47 |
| | S: 25.4 26.8 25.7 25.7 | 4.3 | 0.50 | S: 15.4 15.6 19.9 17.9 | -13.9 | 0.49 | S: 27.1 27.4 32.8 28.1 | -6.8 | 0.49 |
| | E: 14.7 9.1 98.5 25.2 | 83.4 | **0.69** | E: 27.3 18.7 100.0 93.5 | 53.5 | **0.63** | E: 31.0 24.0 100.0 90.9 | 51.3 | **0.62** |
| | M: 17.4 8.7 9.7 90.5 | 86.8 | **0.69** | M: 41.6 12.8 95.8 99.7 | 49.8 | **0.62** | M: 43.8 14.9 93.8 99.7 | 49.0 | **0.62** |
| Cow | P: 14.1 18.8 11.9 17.9 | -15.1 | 0.49 | P: 35.3 34.4 20.2 39.1 | 11.4 | **0.51** | P: 46.5 44.0 31.3 47.3 | 12.3 | **0.51** |
| | S: 16.1 38.3 12.5 27.1 | 51.4 | **0.54** | S: 18.6 62.7 10.7 33.5 | 66.7 | **0.60** | S: 12.8 40.1 12.1 23.6 | 59.7 | **0.56** |
| | E: 2.3 2.3 29.8 2.0 | 92.6 | **0.57** | E: 4.4 8.3 91.9 4.1 | 93.9 | **0.70** | E: 7.2 7.7 90.1 5.3 | 92.5 | **0.70** |
| | M: 8.6 17.4 2.4 52.5 | 82.0 | **0.61** | M: 13.8 46.7 9.3 97.2 | 76.0 | **0.68** | M: 14.1 39.4 10.4 97.7 | 78.2 | **0.68** |
| Cow - fixed negatives | P: 14.1 16.2 12.0 18.6 | -10.7 | 0.49 | P: 35.3 32.7 16.9 46.6 | 9.1 | 0.50 | P: 46.6 39.1 26.4 48.9 | 18.2 | **0.52** |
| | S: 18.7 38.3 17.0 35.8 | -37.8 | **0.53** | S: 38.5 62.7 21.0 69.5 | 31.4 | **0.54** | S: 25.7 40.1 12.8 43.3 | 31.9 | **0.53** |
| | E: 23.5 22.2 29.8 14.0 | 33.3 | **0.52** | E: 6.1 7.9 91.9 4.6 | 93.2 | **0.70** | E: 11.2 9.0 90.1 11.0 | 88.4 | **0.69** |
| | M: 8.9 9.4 2.0 52.5 | 87.1 | **0.61** | M: 65.2 62.6 51.5 97.2 | 38.5 | **0.59** | M: 69.4 55.1 47.4 97.7 | 41.3 | **0.59** |

**Table 1** Binary cross-dataset generalization for two example categories, car and cow. Each matrix contains the object classification performance (AP) when training on one dataset (rows) and testing on another (columns). The diagonal elements correspond to the self results, *i.e.* training and testing on the same dataset. The percentage difference between the self results and the average of the other results per row corresponds to the value indicated in the column "% Drop". CD is our newly proposed cross-dataset measure. We report in bold the values higher than 0.5. P,S,E,M stand respectively for the datasets Pascal VOC07, SUN, ETH80, MSRCORID.

The information obtained in this way over the whole collections provides us only with a small part of the full picture about the bias problem. The dataset recognition performance does not give an insight on how the classes in each collection are related among each other, nor how a model defined for a class in one dataset will generalize to the others. We look into this problem in the following paragraph.

### 4.2 Cross-dataset generalization test

With a procedure similar to that in [39], we perform a cross-dataset generalization experiment over two specific object classes shared among multiple datasets: *car* and *cow*. For the class car we selected randomly from four collections of the sparse set two groups of 50 positive and 1000 negative examples respectively for training and testing. For the class cow we considered 30 positive/1000 negative examples in training and 18 positive/1000 negative examples in testing (limited by the number of cow images in Sun). We repeat the sample selection 10 times and the obtained average precision results are presented in the matrices of Table 1.

Coherently with what deduced from the *name the dataset* experiment, scene-centric (PascalVOC07, SUN) and object-centric (ETH80, MSRCORID) collections appear separated among each other. For the first ones, the low in-dataset results are mainly due to their multi-label nature: an image labelled as people may still contain a car and this creates confusion both at training and at test time. The final effect is that the object-centric collections, with better defined positive and negative sets, provide higher cross-dataset results than the respective $Self$ performance. This becomes even more evident when using DeCAF than with BOWsift.

From Figure 1 we know that annotating the samples of the object-centric datasets with the corresponding collection name is an easy task with almost no confusion between ETH80 and MSRCORID. However, when focusing on the specific results for car and cow, we observe different trends. Learning a *car* model from images of toys or of real objects does not seem so different in terms of the final testing performance when using DeCAF. The diagonal matrix values that were prominent with BOWsift are now surrounded by high average precision results. On the other hand, recognizing a living non-rigid object like a *cow* is more challenging due to its large in-class variability. DeCAF features provide a high performance inside each collection, but the difference between the in-dataset and cross-dataset results remains large almost as with BOWsift. We also re-run the experiments on the class cow by using a fixed negative set in the test always extracted from the training collection. The visible increase in the cross-dataset results indicate that the negative set bias maintain its effect regardless of the used representation.

From the values of $\%Drop$ and $CD$ we see that these two measures may have a different behavior: for the class cow with BOWsift, the $\%Drop$ value for ETH80 (92.6) is higher than the corresponding value for MSRCORID (82.0), but the opposite happens for $CD$ (respectively 0.57 and 0.61). The reason is that $CD$ integrates the information on the in-dataset recognition which is higher and more reliable for MSRCORID. Moreover we also notice that passing from BOWsift to DeCAF the $CD$ value increases in some cases indicating a more significant bias.

Our main conclusion from these experiments is that by using DeCAF we cannot expect to fully solve the dataset bias problem. The capture bias appears class-dependent and DeCAF can help overcoming that in some cases (car, ETH80, MSRCORID), but by capturing detailed information from the images, it might also worsen it w.r.t. using less powerful representations. Furthermore, the negative bias persists regardless of the feature used to represent the data.

### 4.3 Undoing the Dataset Bias

We focus here on the method proposed in [25] to overcome the dataset bias. Our aim is to verify its effect when using the DeCAF features. The *Unbias* approach has a formulation similar to multi-task learning: the available images of multiple datasets are kept separated as belonging to different tasks and a max-margin model is learned from the information shared over all of them.

### 4.3.1 Method.

Let us assume to have $i = 1, \ldots, n$ datasets $D_1, \ldots, D_n$ each consisting of $s_i$ training samples $D_i = \{(\mathbf{x}_1^i, y_1^i), \ldots, (\mathbf{x}_{s_i}^i, y_{s_i}^i)\}$. Here $\mathbf{x}_j^i \in \mathbb{R}^m$ represents the $m$-dimensional feature vector and $y_j^i \in \{-1, 1\}$ represents the label for the example $j$ of dataset $D_i$. Specifically all the datasets share an object class and its images are annotated with label 1, while all the other samples in the different datasets have label -1. The *Unbias* algorithm [25] consists in learning a binary model per dataset $\mathbf{w}_i = \mathbf{w}_{vw} + \Delta_i$, where $\mathbf{w}_{vw}$ is a model for the visual world, while $\Delta_i$ is the bias for each dataset. These two parts are obtained by solving the following optimization problem:

$$\min_{\substack{\mathbf{w}_{vw}, \Delta_i \\ \xi, \rho}} \frac{1}{2}\|\mathbf{w}_{vw}\|^2 + \frac{\lambda}{2}\sum_{i=1}^{n}\|\Delta_i\|^2 + C_1 \sum_{i=1}^{n}\sum_{j=1}^{s_i} \xi_j^i + C_2 \sum_{i=1}^{n}\sum_{j=1}^{s_i} \rho_j^i \tag{1}$$

$$\text{subject to } \mathbf{w}_i = \mathbf{w}_{vw} + \Delta_i \tag{2}$$

$$y_j^i \mathbf{w}_{vw} \cdot \mathbf{x_j^i} \geq 1 - \xi_j^i \tag{3}$$

$$y_j^i \mathbf{w}_i \cdot \mathbf{x_j^i} \geq 1 - \rho_j^i \tag{4}$$

$$\xi_j^i \geq 0, \ \rho_j^i \geq 0 \tag{5}$$

$$i = 1, \ldots, n \ \ j = 1, \ldots, s_i \ . \tag{6}$$

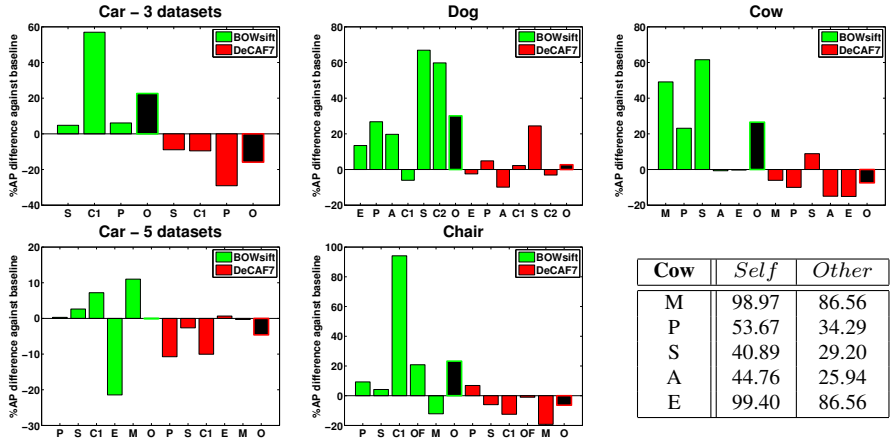

**Fig. 2** Percentage difference in average precision between the results of *Unbias* and the baseline *All* over each target dataset. P,S,E,M,A,C1,C2,OF stand respectively for the datasets Pascal VOC07, SUN, ETH80, MSRCORID, AwA, Caltech101, Caltech256 and Office. With O we indicate the overall value, *i.e.* the average of the percentage difference over all the considered datasets (shown in black).

*4.3.2 Experiments.*

We run the experiments focusing on the classes *car*, *cow*, *dog* and *chair* and reproducing a similar setup to what previously used in [25]. For the class car we consider two settings with three and five datasets, while we use five datasets for cow and chair and six datasets for dog. One of the dataset is left out in round for testing while all the others are used as sources of training samples. We used the original implementation provided by the authors and we ran a preliminary cross-validation to choose the parameters. Specifically, one of the available datasets is always considered in round as target. The cross validation is executed on the remaining source collections with the following parameter ranges: $\lambda = [0.5, 1, 5, 10]$, $C_1 = [10^2, 10^3, 10^4]$, $C_2 = [10, 20, 40, 60, 80, 100]$. The best parameter combination is chosen as the one for which $\mathbf{w}_{vw}$ produces the best result on the source collections and it is then used to obtain the results on the target dataset.

We showed results on four object classes, *car*, *dog*, *chair* and *cow*. To explain the setup we focus here on the first class and the experiment involving three datasets: SUN, Caltech101 and Pascal VOC07. For each of the collections extracted from the sparse setup we identified and separated the positive images belonging to class *car* and the negative images belonging to all the other classes resulting in 777/10106 positive/negative images for SUN, 123/5422 for Caltech101 and 1434/10785 for Pascal VOC07. We divided each positive and negative set into two halves: one was used for training and the other for validation and test. The setup is analogous for all the other experiments on different classes and collections.

We compare the results obtained with the *Unbias* model against those produced by a linear SVM when *All* the training images of the source datasets are considered together. We show the percentage relative difference in terms of average precision for these two learning strategies in Figure 2. The results indicate that, in most cases when using BOWsift the *Unbias* method improves over the plain *All* SVM, while the opposite happens when using DeCAF7. As already pointed out in the previous section, a highly accurate representation may enhance the cross-dataset differences by decreasing the amount of shared information among different collections and the effectiveness of methods that leverage over it. Nevertheless, removing the dataset separation and considering all the images together provides a better coverage of the object variability and allows for a higher cross-dataset performance.

In the last row of Figure 2 we present the results obtained with the class *cow* (on the left) together with the average precision per dataset (on the right) when using DeCAF7. Specifically, the table allows to compare the performance of training and testing on the same dataset ($Self$) against the best result between *Unbias* and *All* (indicated as $Other$). Despite the good performance obtained by directly learning on other datasets, the obtained results are still lower than what can be expected having access to trained samples of each collection. This suggests that, if the final goal is to solve a particular task, an adaptation process from generic to specific may still be needed to close the gap. Similar trends can be observed for the other categories.

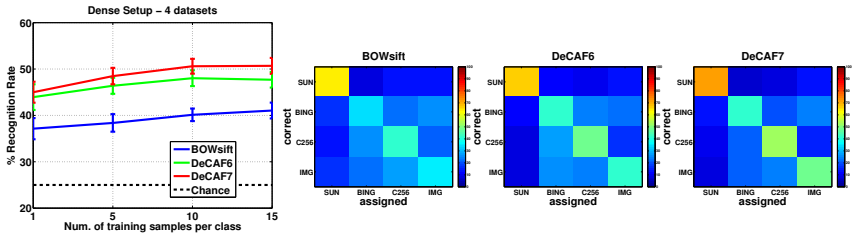

**Fig. 3** Name the dataset experiment over the dense setup with 4 datasets. We use a linear SVM classifier with C value tuned by 5-fold cross validation on the training set, we show average and standard deviation results over 10 repetitions. The title of each confusion matrix indicates the feature used for the corresponding experiments.

|  | BOWsift | | | % Drop | CD | DeCAF7 | | | % Drop | CD |
|---|---|---|---|---|---|---|---|---|---|---|
| | C256 | IMG test | SUN | | | C256 | IMG test | SUN | | |
| C256 | 25.15 | 15.05 | 9.35 | 51.5 | 0.53 | 73.15 | 56.05 | 20.20 | 47.9 | 0.58 |
| IMG | 14.50 | 17.85 | 9.05 | 34.0 | 0.52 | 64.10 | 64.90 | 22.65 | 33.2 | 0.55 |
| SUN | 7.70 | 8.00 | 13.55 | 42.1 | 0.51 | 21.35 | 23.15 | 30.05 | 25.9 | 0.52 |

**Table 2** Multiclass cross-dataset generalization performance (recognition rate). The percentage difference between the self results and the average of the other results per row correspond to the value indicated in the column % $Drop$. CD is our newly proposed cross-dataset measure.

## 5 Studying the Dense set

### 5.1 Name the Dataset

A second group of experiments on the dense setup allows us to analyze the differences among the datasets avoiding the negative set bias. We run again the *name the dataset* test maintaining the balance among the 40 classes shared by Caltech256, Bing, SUN and Imagenet. We consider a set of 5 samples per object class in testing and an increasing amount of training samples per class from 1 to 15. The results in Figure 3 indicate again the better performance of DeCAF7 over DeCAF6 and BOWsift.

From the confusion matrices it is clear that the separation between object- (Bing, Caltech256, Imagenet) and scene-centric (SUN) datasets is quite easy regardless of the representation, while the differences among the object-centric collections become more evident when passing from BOW to DeCAF. We can get a more concrete idea of the DeCAF performance by looking at Figure 4. Here the first two rows present Imagenet images that have been assigned to Caltech256 with BOWsift but which are correctly recognized with DeCAF7. The two bottom rows contain instead Caltech256 images wrongly annotated as Imagenet samples by BOWsift but correctly labelled with DeCAF7. Considering the white background and standard pose that characterize Caltech256 images, together with the less stereotypical content of Imagenet data, the mistakes of BOWsift can be visually justified, nevertheless the DeCAF features overcome them.

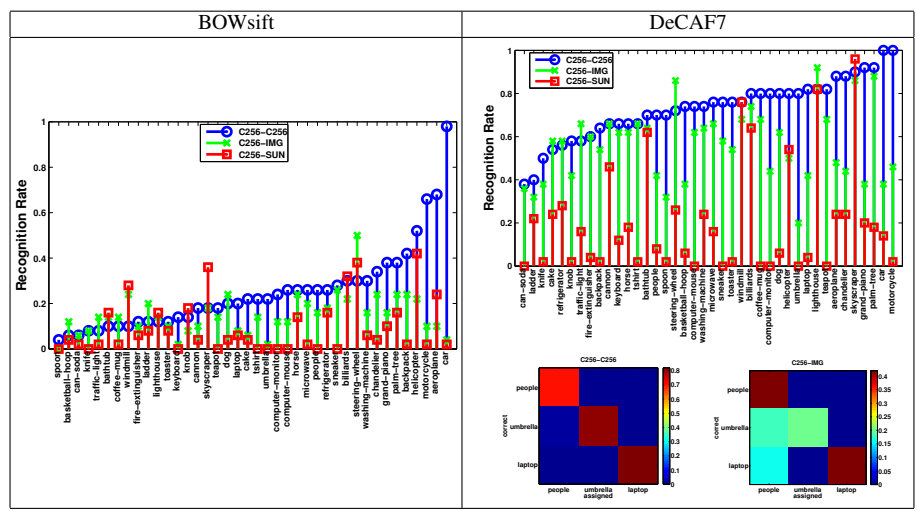

**Table 3** Recognition rate per class from the multiclass cross-dataset generalization test. C256, IMG and SUN stand respectively for Caltech256, Imagenet and SUN datasets. We indicate with "train-test" the pair of datasets used in training and testing.

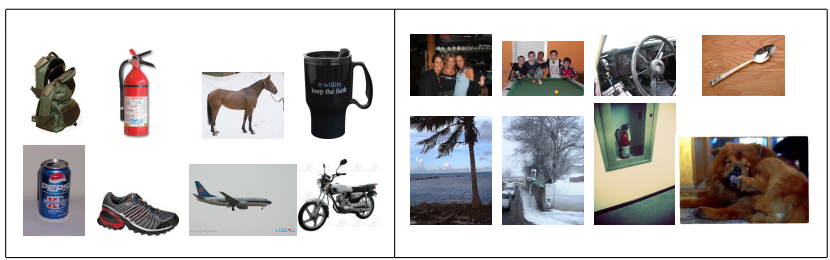

**Table 4** Left: Imagenet images annotated as Caltech256 data with BOWsift but correctly recognized with decaf7. Right: Caltech256 images annotated as Imagenet by BOWsift but correctly recognized with De-CAF7.

Since all the datasets contain the same object classes, we are in fact reproducing a setup generally adopted for domain adaptation [16, 12]. By simplifying the dataset bias problem and identifying each dataset with a domain, we can interpret the results of this experiment as an indication of the domain divergence [2] and deduce that a model trained on SUN will perform poorly on the object-centric collections and vice versa. On the other hand, a better cross dataset generalization should be observed among Imagenet, Caltech256 and Bing. We verify it in the following sections.

5.2 Cross-dataset generalization test.

We consider the same setup used before with 15 samples per class from each collection in training and 5 samples per class in test. However, now we train a one-vs-all multiclass SVM per dataset. Due to its noisy nature we exclude Bing here and we dedicate more attention to it in the next paragraph.

The average recognition rate results over 10 data splits are reported in Table 2. By comparing the values of $\%Drop$ and $CD$ we can observe that they provide opposite messages. The first suggests that we get a better generalization when passing from BOWsift to DeCAF7. However, considering the higher $Self$ result, $CD$ evaluates the dataset bias as more significant when using DeCAF7. The expectation indicated before on the cross-dataset performance are confirmed here: the classification models learned on Caltech256 and Imagenet have low recognition rate on SUN. Generalizing between Caltech256 and Imagenet, instead, appears easier and the results show a particular behavior: although the classifier on Caltech256 tends to fail more on Imagenet than on itself, when training on Imagenet the in-dataset and cross-dataset performance are almost the same. Of course we have to remind that the DeCAF features where defined over Imagenet samples and this can be part of the cause of the observed asymmetric results.

To visualize the effect of the dataset-bias per class we present the separate recognition rate in Table 3. Specifically we consider the case of training on Caltech256. From the top plot we can see that *motorcycle*, *aeroplane* and *car* are the objects better recognized when testing on Caltech256 with BOWsift and they are also the classes that mostly contribute to the recognition drop when testing on Imagenet and SUN. On the other hand the classes *steering-wheel*, *windmill*, *bathtub*, *lighthouse* and *skyscraper* are better recognized on SUN and/or Imagenet than on Caltech256. All these last objects occupy most part of the image in all the collections and present less dataset-specific characteristics. When looking at the results with DeCAF7, *motorcycle* and *car* are still among the classes with the highest cross-dataset recognition difference, together with *people*, *spoon*, *umbrella*, *basketball-hoop* and *laptop*.

As already indicated by the binary experiments, even these results confirm that the dataset bias is in fact class dependent and that using DeCAF does not automatically solve the problem. A further remark can be done here about Imagenet. Although often considered as one of the less biased collections it actually presents a specific characteristic: the images are annotated with a single label but in fact may contain more than one visual category. In particular, its images often depict people even when they are labelled with a different class name. As a demonstration we report at the bottom of Table 3 a sub-part of the confusion matrix when training on Caltech256 and testing both on itself and on Imagenet. The results show that people are recognized in the class umbrella and laptop with relevant influence on the overall annotation errors.

### 5.3 Noisy Source Data and Domain Adaptation.

Until now we have discussed and demonstrated empirically that the difference among two data collections can actually originate from multiple and often co-occurring causes. However the standard assumption is that the label assigned to each image is correct. In some practical cases this condition does not hold and it happens that the available set of annotated data on which a model can be trained is noisy. This is the typical condition of learning from web data where the CNN features have greatly demonstrated their potential against handcrafted descriptors [5]. In this setting the noisiness is typically due to the way in which the images are gathered. A textual web search output contains not only images that cover several meanings of the queried word, but also unrelated images not containing the searched visual concept.

To overcome this kind of bias one possible strategy is to group or subselect the data while training a model on them. Some domain adaptation strategies seem perfectly suited for this task. The *landmark* method introduced in [13] allows to sample the source training images that are more relevant for the target test data. This approach, as well as the Geodesic Flow Kernel [16] on which it builds, and the related *Subspace Alignment* (SA) method [12] are all feature-based adaptive approaches that have been previously tested mainly on the Office dataset. Another relevant technique is the *reshape* approach proposed in [22] that was previously used on Bing images to identify latent sub-domains. In the following we briefly review these methods and then we evaluated them for cross-dataset generalization when training on Bing and testing both on Caltech256 and SUN.

### 5.3.1 Methods

*Landmark.* The landmark method [13] was explicitly indicated as a possible strategy to overcome dataset bias in [15]. Let $D_s = \{(\mathbf{x}_m, y_m)\}_{m=1}^M$ denote data points and their labels from the source domain and likewise $D_t = \{\mathbf{x}_n\}_{n=1}^N$ for the target domain. An indicator variable $\alpha_m \in \{0, 1\}$ is assigned to each source sample and identified by minimizing the Maximum Mean Discrepancy [18]:

$$\min_\alpha \left\| \frac{1}{\sum_m \alpha_m} \sum_m \alpha_m \phi(\mathbf{x}_m) - \frac{1}{N} \sum_n \phi(\mathbf{x}_n) \right\|^2 \tag{7}$$

$$\text{s.t.} \quad \frac{1}{\sum_m \alpha_m} \sum_m \alpha_m y_{mc} = \frac{1}{M} \sum_m y_{mc} \tag{8}$$

where $\phi(\mathbf{x})$ is a kernel mapping function. Here the constraint is added to have the same class statistics in the selected landmarks as in the original data. The optimization is solved introducing the variables $\beta_m = \frac{\alpha_m}{\sum_m \alpha_m}$ and relaxing them to live in the simplex $\{0 \leq \beta_m \leq 1, \sum_m \beta_m = 1\}$. The binary solution for $\alpha_m$ is recovered by thresholding $\beta_m$. The geodesic flow kernel (GFK) [16], computed between the source $D_s$ and the target $D_t$, is used to compose the kernel mapping function $\phi(\mathbf{x})$:

$$\phi(\mathbf{x}_i)^\top \phi(\mathbf{x}_j) = K(\mathbf{x}_i, \mathbf{x}_j) \tag{9}$$
$$= \exp\{-(\mathbf{x}_i - \mathbf{x}_j)^\top G(\mathbf{x}_i - \mathbf{x}_j)/\sigma^2\} \tag{10}$$

For different values of the bandwidth $\sigma_q$ the corresponding landmarks $\mathcal{L}^q$ are identified and a new domain pair is obtained with source $D_s \setminus \mathcal{L}^q$ and target $D_t \cup \mathcal{L}^q$. The samples similarities over all the auxiliary domain pairs are integrated using multiple kernel learning.

This method has several parameters. First of all GFK needs the definition of a subspace dimensionality. For our experiments we adopted the Subspace Disagreement (SD) measure [16] to choose this dimensionality. We used the code for the *landmark* method provided by the authors considering $\sigma_q = 2^q \sigma_0$ with $\sigma_0$ equal to the median distance over all the pairwise data points and $q = [-2, -1, 0, 1, 2]$. The threshold to choose the source samples is fixed as the median of all the $\beta_m$ values, and the $C$ parameter for the multiple kernel learning algorithm is chosen in the range $[10^{-1}, 10^0, 10^1, 10^2, 10^3, 10^4]$, exploiting the original model selection that automatically considers $D_s \setminus \sum_q \mathcal{L}^q$ as a validation set.

*Subspace Alignment (SA).* This approach learns an alignment matrix $A$ between the source and the target subspace by minimizing the following Bregman divergence [12]:

$$\|X_s A - X_t\|_F^2 . \tag{11}$$

Here $X_s$ and $X_t$ are the subspace bases obtained by applying PCA over the source and target data and selecting for each domain the $d$ eigenvectors corresponding to the $d$ largest eigenvalues. Specifically we set $d$ equal to the SD measure [16] and we run the experiments with *SA* using the code provided by the authors. The $C$ parameter is automatically chosen as the one that produces the best recognition on the source set in the range $[10^{-6}, 10^{-5}, \ldots, 10^3]$.

*Domain Adaptation Machine (DAM).* The DAM method [8] was originally developed for the transfer learning setup where at least few target training labelled samples are available, but it can also be applied in the unsupervised setting [41]. Specifically, DAM is formulated as a regression task solving the following optimization problem:

$$\min_{\mathbf{f^t}, \mathbf{w}, b, \xi_i, \xi_i^*} \frac{1}{2}\|\mathbf{w}\|^2 + C \sum_{i=1}^{n_t} (\xi_i + \xi_i^*)$$
$$+ \frac{1}{2}\theta \left( \|\mathbf{f}_l^t - \mathbf{y}_l\|^2 + \sum_s \gamma_s \|\mathbf{f}_u^t - \mathbf{f}_u^s\|^2 \right) \tag{12}$$

$$\text{s.t.} \quad \mathbf{w} \cdot \phi(x_i) + b - f_i^t \leq \epsilon + \xi_i, \quad \xi_i \geq 0, \tag{13}$$

$$f_i^t - \mathbf{w} \cdot \phi(x_i) + b \leq \epsilon + \xi_i^*, \quad \xi_i^* \geq 0 \tag{14}$$

where $\xi_i, \xi_i^*$ are slack variables for the $\epsilon$-insensitive loss. With $\mathbf{f}_l^t$ we indicate the predictions of the learned model over the labelled target samples and $\mathbf{y}_l$ are the corresponding ground truth labels. $\mathbf{f}_u^t$ and $\mathbf{f}_u^s$ are respectively the predictions of the target model and of the source model $s$ on the unlabelled part of the target. In the unsupervised case the first term in the parentheses of equation (12) is absent.

We used the implementation provided by the authors fixing the parameter for a single source as in the original code: $\theta = 1$ and $\gamma_1 = 0.5$. The choice of the $\gamma_s$ parameters in case of multiple sources is discussed in the following paragraph. The source models are trained with a linear SVM choosing the $C$ parameter by two-fold cross validation on the sources in the range $[10^{-6}, 10^{-5}, \ldots, 10^3]$. For DAM we adopted the $C$ value that was producing the best average result on the sources.

*Discovering Latent Domains (reshape).* Each dataset may contain multiple domains. To discover them, the method proposed in [22] considers two main steps. If there are $K$ semantic categories and $S$ domains the data points are grouped into $J = K \times S$ local clusters. Intuitively each local cluster will contain the data points from one domain and one semantic category. In the second stage the means of the local clusters are grouped in order to identify the domains clusters. Two constraints must be respected: each local cluster can contain only data points of a single class and each domain cluster can contain only one local cluster from each object category. The optimization problem is formulated as an EM iteration algorithm with the two stages repeated until convergence.

This method have already been used to separate the Bing images in sub-domains [22]. In our work we applied it on the $K = 40$ categories of the dense setup and we searched for $S = 2$ domains. For each of the two domains we considered the combination with the *SA* and *DAM* methods.

*reshape + SA:* each of the two obtained domain is used separately as source set and the SA method is applied to obtain the alignment with the target set. We collected the results for both the sources over all the experimental splits and we selected only the best results corresponding to the source producing the highest recognition rate on the target. This setup is chosen to analyze the ideal condition of an oracle providing the target labels. Our goal is to evaluate if, in this best-case scenario, it would be possible see an improvement over the case where all the source samples are considered at once.

*reshape + DAM:* we followed the same logic even when combining the discovered sub-domains with DAM. In this case the sources are not used separately but together trying all the possible combinations with $\gamma_1 = [0.1 : 0.1 : 1]$ and $\gamma_2 = 1 - \gamma_1$.

*Self-Labelling.* A simple strategy for domain adaptation consists in subselecting and using the target samples while learning the source model. This technique is known as *self-labelling* [3, 20] and starts by annotating the target with a classifier trained on the source. The target samples for which the source model presents the highest confidence are then used together with the source samples in the following iteration. In our experiments a one-vs-all SVM model is trained on the source data with the C parameter chosen by cross validation. At each iteration the model is used to classify on the target data and the images assigned to every class are ranked on the basis of their output margin. Only the images with a margin higher than the average are selected and sorted by the difference between the first and the second higher margin over the classes. The top samples in the obtained list per class are then used in training with the pseudo-labels assigned to them in the previous iteration. We set the number of iterations to 10 and the number of selected target samples per class to 2.

### 5.4 Experiments

For our experiments we consider an increasing number of training images per class from 10 to 50 and we test on 30 images per class on Caltech256 and 20 images per class on SUN. The experiments are repeated on 10 random data splits.

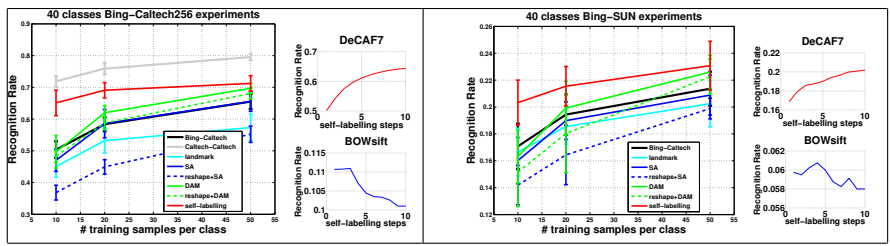

**Fig. 4** Results of the Bing-Caltech256 and Bing-SUN experiments with DeCAF7. We report the performance of different domain adaptation methods (big plots) together with the recognition rate obtained in 10 subsequent steps of the self-labelling procedure (small plots). For the last ones we show the performance obtained both with DeCAF7 and and with BOWsift when having originally 10 samples per class from Bing.

In Figure 4 we present the results obtained using DECAF7 with the landmark method, and the reshape approach combined respectively with SA and Domain Adaptation Machine (DAM [8]). More in details, we use the reshape method to divide the Bing images into two subgroups and we apply SA to adapt each of them to the target. We identify the best subgroup as that with the highest target performance and we report its results. The subgroups are instead considered as two source domains with DAM: we assign different weights to their relative importance and show also in this case the best obtained performance[4]. As reference we also present the performance of the SA and DAM method without reshaping. Finally we test a simple *self-labelling* strategy that was already used in [38]. Differently from the previously described techniques this method learns a model over the full set of Bing data and progressively selects target samples to augment the training set.

The obtained results go in the same direction of what observed previously with the *Unbias* method. Despite the presence of noisy data, subselecting them (with landmark) or grouping the samples (reshape+SA, reshape+DAM) do not seem to work better than just using all the source data at once. On the other way round *self-labelling* [38] consistently improves the original results with a significant gain in performance especially when only a reduced set of training images per class is available. One well known drawback of this strategy is that subsequent errors in the target annotations may lead to significant drift from the correct solution. However, when working with DeCAF features this risk appears highly reduced: this can be appreciated looking at the recognition rate obtained over ten iterations of the target selection procedure, considering in particular the comparison against the corresponding performance obtained when using BOWsift (see the small plots in Figure 4).

## 6 Conclusions

In this paper we attempted at positioning the dataset bias problem in the CNN-based features arena with an extensive experimental evaluation. At the same time, we pushed the envelope in terms of the scale and complexity of the evaluation proto-col, so to be able to analyze all the different nuances of the problem. We focused on DeCAF features, as they are the most popular CNN-learned descriptors, and for the impressive results obtained so far in several visual recognition domains.

---

[4] More details on the experimental setup can be found in the supplementary material.

A first main result of our analysis is that DeCAF not only does not solve the dataset bias problem in general, but in some cases (both class- and dataset-dependent) they capture specific information that induce worse performance than what obtained with less powerful features like BOWsift. Moreover, the negative bias remains, as it cannot intrinsically be removed (or alleviated) by changing feature representation. A second result concerns the effectiveness of learning methods applied over the chosen features: nor a method specifically designed to undo the dataset bias, neither algorithms successfully used in the domain adaptation setting seem to work when applied over DeCAF features. It appears as if the highly descriptive power of the features, that determined much of their successes so far, in this particular setting backfire, as it makes the task of learning how to extract general information across different data collection more difficult. Interestingly, a simple selection procedure based on self labeling over the test set leads to a significant increase in performance. This questions whether methods effectively used in domain adaptation should be considered automatically as suitable for dataset bias, and vice versa. How to leverage over the power of deep learning methods to attack this problem in all its complexity, well represented by our proposed experimental setup, is open for research in future work.

A Deeper Look at Dataset Bias    1943. Xu, Z., Li, W., Niu, L., Xu, D.: Exploiting low-rank structure from latent domains for domain generalization. In: ECCV (2014)
44. Zeiler, M.D., Fergus, R.: Visualizing and understanding convolutional networks. In: ECCV (2014)
45. Zhang, N., Donahue, J., Girshick, R., Darrell, T.: Part-based R-CNNs for fine-grained category detection. In: ECCV (2014)